# Vision–Language Models for Ergonomic Assessment of Manual Lifting Tasks: Estimating Horizontal and Vertical Hand Distances from RGB Video

[a] Mohammad Sadra Rajabi, https://orcid.org/0000-0002-9100-3973

[b] Aanuoluwapo Ojelade, https://orcid.org/0000-0001-9715-3254

[a] Sunwook Kim, https://orcid.org/0000-0003-3624-1781

[a] Maury A. Nussbaum, https://orcid.org/0000-0002-1887-8431

**Affiliations:**

[a] Department of Industrial and Systems Engineering, Virginia Tech, Blacksburg VA 24061, USA

[b] St. Jude Children's Research Hospital, Memphis, TN 38105, USA

**Corresponding author:**

Corresponding address: Maury A. Nussbaum

Department of Industrial and Systems Engineering,

Virginia Tech, Blacksburg VA 24061, USA.

Phone: 540-231-6053. Email: nussbaum@vt.edu

**Abstract**

Manual lifting tasks are a major contributor to work-related musculoskeletal disorders, and effective ergonomic risk assessment is essential for quantifying physical exposure and informing ergonomic interventions. The Revised NIOSH Lifting Equation (RNLE) is a widely used ergonomic risk assessment tool for lifting tasks that relies on six task variables, including horizontal ($H$) and vertical ($V$) hand distances; such distances are typically obtained through manual measurement or specialized sensing systems and are difficult to use in real-world environments. We evaluated the feasibility of using innovative vision–language models (VLMs) to non-invasively estimate $H$ and $V$ from RGB video streams. Two multi-stage VLM-based pipelines were developed: a text-guided detection-only pipeline and a detection-plus-segmentation pipeline. Both pipelines used text-guided localization of task-relevant regions of interest, visual feature extraction from those regions, and transformer-based temporal regression to estimate $H$ and $V$ at the start and end of a lift. For a range of lifting tasks, estimation performance was evaluated using leave-one-subject-out validation across the two pipelines and seven camera view conditions. Results varied significantly across pipelines and camera view conditions, with the segmentation-based, multi-view pipeline consistently yielding the smallest errors, achieving mean absolute errors of approximately 6–8 cm when estimating $H$ and 5–8 cm when estimating $V$. Across pipelines and camera view configurations, pixel-level segmentation reduced estimation error by approximately 20–30% for $H$ and 35–40% for $V$ relative to the detection-only pipeline. These findings support the feasibility of VLM-based pipelines for video-based estimation of RNLE distance parameters.

**Keywords**: Vision–Language Models (VLMs); Revised NIOSH Lifting Equation (RNLE); Computer vision; Ergonomic risk assessment; Segmentation; Multi-view video

# 1.0 Introduction

Work-related musculoskeletal disorders (WMSDs) remain a major occupational health concern worldwide and are among the leading causes of lost workdays, reduced productivity, and substantial economic burden for employers and workers (Govaerts et al., 2021; Liberty Mutual Insurance, 2023; U.S. Bureau of Labor Statistics, 2024). Manual material handling (MMH) tasks—such as lifting, lowering, carrying, pushing, and pulling—are a primary contributor to WMSDs, as they expose workers to biomechanical risk factors including forceful exertions, non-neutral postures, repetitive motions, and prolonged task durations (Da Costa & Vieira, 2010; Waters et al., 1993). Epidemiological evidence consistently demonstrates elevated incidence rates of low back and upper-extremity disorders among workers engaged in MMH-intensive occupations across manufacturing, logistics, healthcare, and construction sectors (Punnett & Wegman, 2004). Reducing the burden of WMSDs therefore requires accurate and practical methods for quantifying physical exposures and injury risks during MMH tasks in real-world work environments.

Ergonomic risk assessment tools, such as the Revised NIOSH Lifting Equation (RNLE; Waters et al., 1993), are widely used to evaluate the physical demands of manual lifting tasks and to guide workplace interventions. The RNLE relies on six task-specific parameters—including horizontal and vertical hand distances, load weight, asymmetry, coupling quality, and task frequency—that directly affect biomechanical loading and WMSD risk (Waters et al., 1993). Accurate measurement of these parameters is critical, as measurement errors have been shown to affect RNLE outputs and risk estimates, potentially influencing risk classification and subsequent intervention decisions (Fox et al., 2019; Waters et al., 1998). In practice, though, RNLE parameters are commonly obtained through manual measurement, which is time-consuming, subject to observer bias, and difficult to apply consistently across large or complex work environments (David, 2005; Dempsey et al., 2001, 2005; Lu et al., 2016). Although wearable sensors (Hlucny & Novak, 2020; Lu et al., 2020; Mudiyanselage et al., 2021; Ranavolo et al., 2024) and marker-based motion capture systems (Gutierrez et al., 2024; Ranavolo et al., 2017) can provide accurate measurements, their cost, intrusiveness, and logistical complexity may limit their feasibility for practical field-based ergonomic assessments (Sabino et al., 2024; Schall et al., 2018).

To address these limitations, computer vision–based approaches have been increasingly explored for ergonomic assessment. Markerless human pose estimation methods using RGB or RGB-D video—such as OpenPose (Zhao et al., 2023) or BlazePose (Bazarevsky et al., 2020)—can estimate body joint locations and postures without physical instrumentation (Cao et al., 2016). These approaches have shown promise for posture classification (Chen, 2019; Hsu et al., 2025; Jung et al., 2022; Liu & Chang, 2022) and ergonomic risk assessment (Forgione et al., 2025; Lou et al., 2025; Pires et al., 2025), but their reported application to distance-based ergonomic parameters remains limited. Like most vision-based methods, pose-based systems are sensitive to occlusion, camera perspective, and lighting conditions; additionally, they often produce temporally unstable body joint keypoint trajectories across frames, which can degrade frame-level distance estimates (Cheng et al., 2019; Mehta et al., 2017; Pavllo et al., 2018; Veges & Lorincz, 2020; Zahabi et al., 2026). Moreover, such systems primarily focus on skeletal representations and do not explicitly encode object-centric spatial relationships—such as those between the worker, load, and environment—that are fundamental to ergonomic risk modeling (Bezzini et al., 2023; Parsa et al., 2019; Paudel et al., 2022). As a result, pose-based approaches may struggle to provide robust estimates of the geometric parameters required for ergonomic risk assessment tools such as the RNLE.

Recent advances in *vision–language models* (VLMs) offer clear potential to overcome these challenges by enabling joint reasoning about objects, actions, and spatial relationships (e.g., relative position and distance) between workers, handled loads, and the environment within complex scenes (Guran et al., 2024; Liao et al., 2024; Zhang et al., 2024). VLMs integrate visual encoders with language-based representations, allowing models to be prompted to directly identify and localize semantically meaningful entities—such as the worker, relevant body segments, handled loads, and tools—using natural-language queries, rather than relying solely on skeletal representations (Chen et al., 2024; Kirillov et al., 2023; Liu et al., 2024). This combination of text-guided object localization and object-centric visual representations suggests that VLM-based approaches may be well suited for estimating distance-based ergonomic parameters, since these representations facilitate explicit modeling of human–object and object–environment relationships that are difficult to capture using pose-based methods alone (Kang et al., 2025; Wen et al., 2025). By combining text-guided detection and segmentation with geometric reasoning, VLMs can provide spatially grounded representations of key regions of interest that correspond directly to the definitions of horizontal and vertical distances in the RNLE. However, as visibility of task-relevant landmarks can vary with camera viewpoint, it is important to evaluate the effect of camera view conditions on VLM-based distance estimation. Despite the integration of VLMs in general computer vision and emerging applications in construction safety (Fan, Mei, Wang, et al., 2024) and posture and task classification (Fan, Mei, & Li, 2024; Rajabi et al., 2025; Yong et al., 2024), the use of VLMs for quantitative estimation of ergonomic exposure metrics, particularly RNLE parameters remains unexplored. Furthermore, the effect of camera view conditions on such estimates has not yet been reported to our knowledge.

In the current study, we evaluated the feasibility of using VLMs to estimate the horizontal and vertical distances required by the RNLE from RGB video streams. We examined the estimation performance of VLM-based pipelines that integrate text-guided object detection and segmentation, visual feature extraction, and a transformer-based distance regression model across varied camera view conditions. We hypothesized that estimation performance would vary across VLM-based pipelines and camera view conditions. While our immediate goal was to assess VLM performance under simulated lifting tasks, this work also provides insight into the potential of video-based approaches for estimating key parameters used in ergonomic risk assessment tools across occupational settings.

## 2.0 Methods
### 2.1 Overview of the Dataset

We used a subset of data obtained from a prior study by Ojelade et al. (2025), in which 32 healthy young adults (19 males and 13 females; see Appendix A.1 for participant demographics and inclusion criteria) performed eight manual handling tasks. In the subset of data used here, participants performed symmetric lifts from two distinct *Lift Origins*—the floor and individual knee height—up to individual hip height. Hip height was defined as the vertical distance from the floor to the greater trochanter. Participants completed these lifts using two *Hand Configurations* (broad = 52 cm handle spacing and narrow = 33 cm; see Appendix A.2 for symmetric lifting task details and Figure A.1 for an illustration of the hand configurations), three *Box Mass* levels (6, 9, and 12 kg), with each condition performed twice (2 *Lift Origins* × 2 *Hand Configurations* × 3 *Box Masses* × 2 replications = 24 trials per participant; see Appendix A.3 for experimental procedures).

Whole-body kinematics were recorded using a multimodal instrumentation setup, including three synchronized Azure Kinect™ cameras (Microsoft Corporation, Seattle, WA, USA) and a wearable inertial measurement unit (IMU)-based motion capture (Noraxon Ultium, Noraxon, Scottsdale, AZ, USA). The cameras recorded at 30 Hz and were positioned approximately 1.74 m from edge of the work area (see Appendix A.4 for camera system instrumentation). The three cameras provided distinct viewpoints of the lifting task and are hereafter referred to as Camera View 1 (V1), Camera View 2 (V2), and Camera View 3 (V3). V1 and V2 corresponded to two oblique views of the workspace from opposite sides, whereas V3 provided a frontal view of the participants. To assess the effect of *Camera View Condition*, seven camera view conditions were defined based on the available RGB streams: three single-view conditions (i.e., V1, V2, and V3) and four multi-view conditions formed by combining synchronized views (i.e., V1+V2, V1+V3, V2+V3, and V1+V2+V3). For multi-view conditions, the synchronized views were processed in parallel and combined for model input and evaluation. The IMU system consisted of 16 sensors, positioned on the head, upper and lower thoracic spines, pelvis, and bilaterally on the upper and lower arms, hands, thighs, shanks, and feet; IMU data processing is described in Appendix A.5. For the present study, only the RGB video streams (1280 × 720 pixels) from the Kinects were used as input to the VLM-based pipeline, while the IMU data were used exclusively to derive reference kinematic measurements for ground-truth distance labeling (see Data Labeling section).

### 2.2 Overview of the VLM-Based Pipelines for Horizontal and Vertical Distance Estimation

We developed and evaluated two VLM-based, multi-stage pipelines to estimate RNLE horizontal ($H$) and vertical ($V$) distances from RGB video (Figure 1). The two pipelines are hereafter referred to as GD–Dv2 (Grounding DINO with DINOv2 features; detection-only) and GD–SAM–Dv2 (detection followed by segmentation). For ground-truth labeling, frame-level $H$ and $V$ values were derived from processed IMU-based kinematic data and aligned to RGB video frames. The VLM-based pipelines then consisted of three primary steps: Step 1 was region-of-interest processing, in which task-relevant regions of interest (ROIs) are detected—and, when applicable, segmented—from RGB video frames. Step 2 was feature extraction, in which detected ROIs are transformed into feature representations. Step 3 was distance estimation, in which temporally ordered feature sequences are processed using a transformer-based regression model to estimate $H$ and $V$ at the frame level. Each of these steps is described in detail in the following sections. All processing was performed offline using Python (v3.10.11; https://www.python.org) on secure Virginia Tech Advanced Research Computing resources (VT ARC; https://arc.vt.edu/), with no participant videos, images, or derived data uploaded to or processed on external or cloud-based systems. RGB video handling and frame extraction were implemented using OpenCV (https://opencv.org/), and model inference and training were conducted using PyTorch (https://pytorch.org/, Paszke et al., 2019).

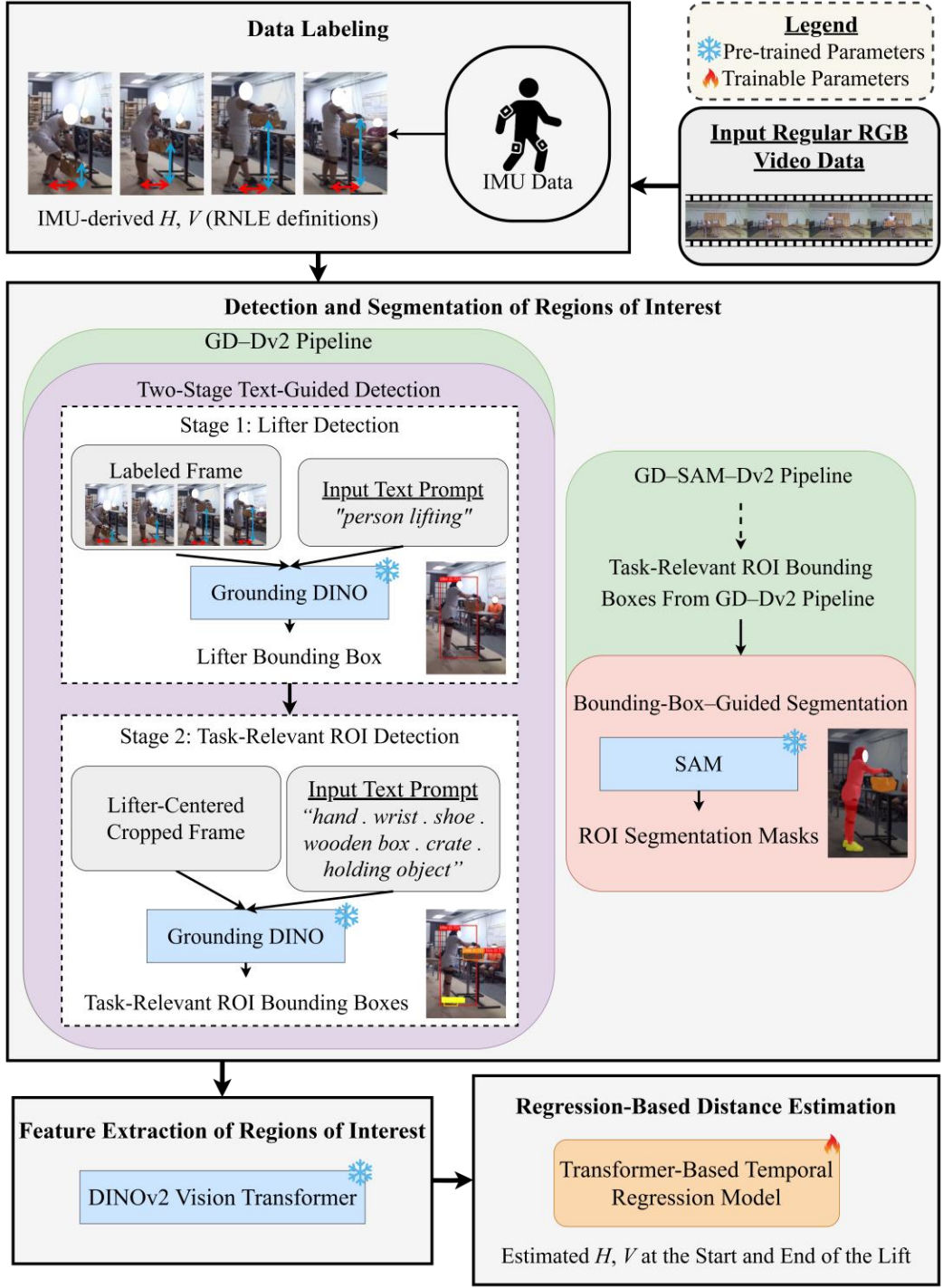

Figure 1: Overview of two pipelines for estimating the horizontal ($H$) and vertical ($V$) distances defined in the RNLE at the start and end of a lift. A text-guided detection process was used to localize the lifter and task-relevant regions of interest (ROIs) using Grounding DINO. In the GD–SAM–Dv2 pipeline, detected ROIs were further refined using bounding-box–guided segmentation with SAM. ROI features extracted using DINOv2 were processed by a transformer-based temporal regression model to estimate $H$ and $V$ at the start and end of a lift.

**2.3 Data Labeling**

As introduced earlier, the RNLE combines six lifting task-related parameters (Appendix A.6), including the horizontal and vertical distances of the hands (i.e., $H$ and $V$; Waters et al., 1993; Figure 2). $H$ is defined as the horizontal distance between the midpoint of the hands and the midpoint of the ankles, projected onto the floor plane (Waters et al., 1993, 1998). Here, for each time sample in the IMU data, the horizontal position of the hands was calculated as the midpoint of the left and right hand-tip trajectories in the anterior–posterior and mediolateral directions, while the horizontal position of the ankles was calculated as the midpoint of the left and right medial malleoli. $H$ was measured based on the Euclidean distance between these two midpoints in the horizontal plane. $V$ is defined as the vertical position of the hands relative to the floor (Waters et al., 1993, 1998). Here, the vertical position of the hands was calculated as the mean of the left and right hand-tip vertical coordinates, since the floor vertical coordinates were set to zero. Each RGB video frame was assigned corresponding ground-truth $H$ and $V$ values by temporally aligning the distances computed from IMU-derived kinematic data with the RGB video recordings at the frame level. These distances (i.e., $H$ and $V$) were used exclusively as reference labels for model training and evaluation.

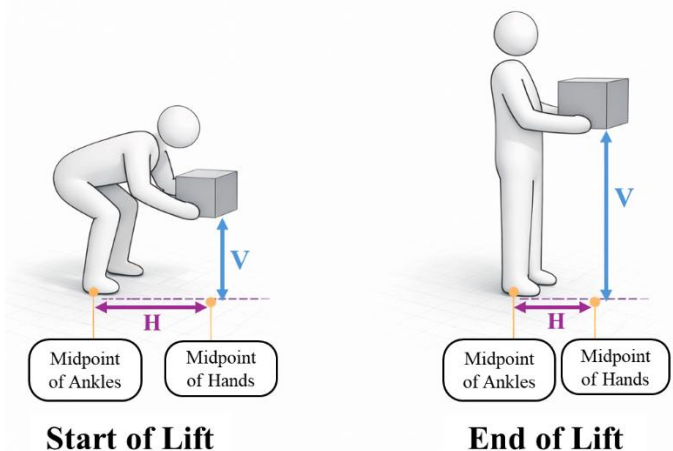

Figure 2: Illustration of the horizontal ($H$) and vertical ($V$) distances defined in the RNLE at the start (left) and end (right) of a lift. Distance $H$ is measured from the midpoint between the ankles to a point projected on the floor directly below the midpoint between the hands, whereas distance $V$ is measured from the floor to the midpoint between the hands (Image generated by ChatGPT, OpenAI, https://openai.com/index/chatgpt/).

**2.4 Detection and Segmentation of Regions of Interest (ROIs)**

To identify task-relevant ROIs required for estimating RNLE distance parameters from RGB video, two VLM–based pipelines were evaluated. Both pipelines relied on text-guided object localization to detect the primary "lifter", as well as relevant body parts and objects, but differed in whether pixel-level segmentation was applied following detection. The two pipelines consisted of: (1) GD–Dv2, a Grounding DINO–based detection-only pipeline (Liu et al., 2024); and (2) GD–SAM–Dv2, a Grounding DINO–based detection followed by segmentation using the Segment Anything Model (SAM: ViT-H backbone; Kirillov et al., 2023). Grounding DINO was used for zero-shot, text-guided object detection, whereas SAM provided zero-shot, promptable segmentation to refine detected ROIs at the pixel level, with neither model requiring task-specific retraining (see Figures 3, A.2, and A.3 for examples of the ROI detection and segmentation process across camera views).

In both pipelines, a two-stage detection strategy was applied for each RGB video frame. First, the "lifter" was identified using the text prompt "*person lifting*", and the detection with the highest confidence score was selected when multiple candidates were present. Second, a spatial crop, centered on the detected lifter, was generated to reduce background interference and to improve localization of fine-grained elements. Within this cropped region, task-relevant details were detected using the prompt "*hand . wrist . shoe . wooden box . crate . holding object*", allowing detection and localization of ROIs. These prompts were selected to align with RNLE-relevant anatomical landmarks and task objects, which was intended to support zero-shot detection.

In the GD–Dv2 pipeline, the resulting bounding boxes from this detection process were used directly as ROIs for subsequent feature extraction and distance estimation. This approach relied solely on box-level localization to represent task-relevant regions. In the GD–SAM–Dv2 pipeline, detected bounding boxes were further refined using the SAM (Kirillov et al., 2023). For each detected ROI, SAM generated a pixel-level segmentation mask guided by the bounding box generated by Grounding DINO (Liu et al., 2024). These masks were used to isolate the detected ROIs from the surrounding background, excluding background pixels and more precisely capturing the shape and spatial extent of each ROI. The segmented regions were then used for feature extraction.

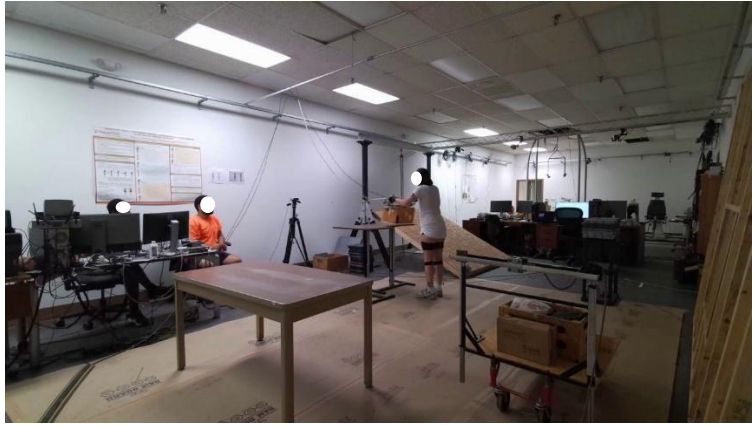 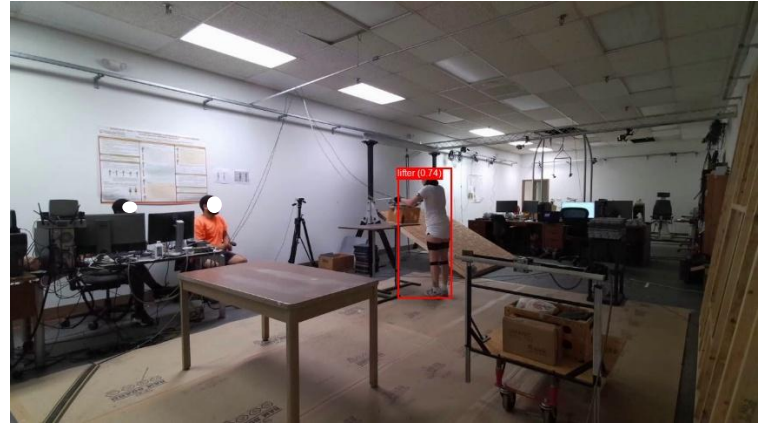
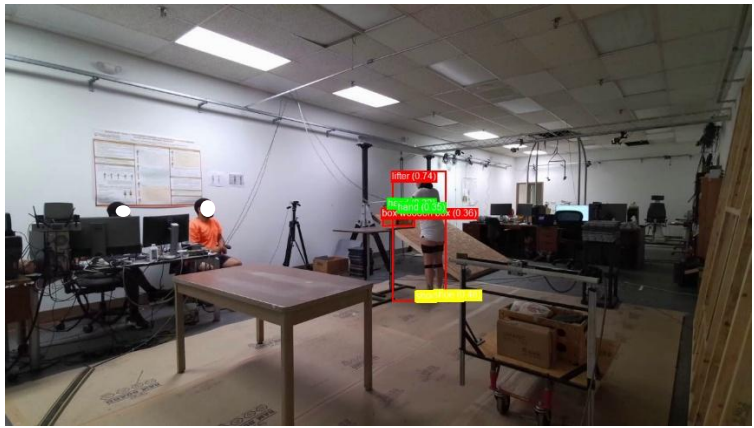 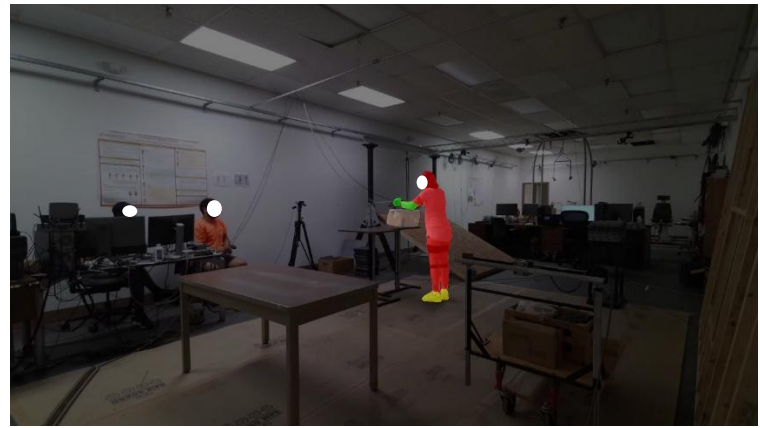

Figure 3: Example visualization of the region-of-interest (ROI) detection and segmentation process for a representative lifting frame of Camera View 1 (V1). Top left: Original RGB video frame. Top right: Detection of the primary lifter using Grounding DINO, with the bounding box corresponding to the highest-confidence person detection. Bottom left: Detection of task-relevant ROIs within a cropped region centered on the lifter using Grounding DINO. Bottom right: Pixel-level segmentation of the detected ROIs produced by the GD–SAM–Dv2 pipeline.

### 2.5 Feature Extraction of Regions of Interest

For each RGB video frame, each ROI was cropped from the original RGB frame based on the detected bounding box or segmentation mask, resized to 224 × 224 pixels, and then normalized using ImageNet statistics (Deng et al., 2009). ImageNet channel-wise statistics were used to normalize RGB pixel values, which standardizes input intensity distributions across frames and ensures compatibility with the pretrained feature extractor (Deng et al., 2009). Visual features were extracted from all identified ROIs using the DINOv2 vision transformer (ViT-Base; Oquab et al., 2024) in a zero-shot configuration without task-specific fine-tuning. DINOv2 is a self-supervised vision transformer trained to produce general-purpose visual representations without reliance on labeled training data (Oquab et al., 2024). Feature extraction was performed identically for both pipelines, to ensure that any observed differences in performance were attributable to differences in ROI representation rather than feature encoding. The resulting 768-dimensional DINOv2 feature vector was extracted for each ROI.

To incorporate task-relevant geometric context, DINOv2 visual features were augmented with object-related geometric features derived from the handled box used in the lifting task. Specifically, five additional features were appended to the DINOv2 feature vector: the width and height of the handled-object bounding box produced by Grounding DINO, normalized by the original frame width and height, respectively, and the known physical dimensions (width, depth, and height) of the box. The normalized image-plane dimensions were computed consistently in both pipelines and captured the apparent size within each frame, while the known physical dimensions provided a consistent real-world scale reference across frames. This reference was necessary to relate image-based representations to physical distances, given that monocular RGB images lack absolute scale information without an object of known size (Jüngel et al., 2008). Together, these features resulted in a 773-dimensional representation for each frame, consisting of DINOv2 visual features augmented with object-related geometric information.

### 2.6 Regression-Based Distance Estimation

Regression-based models were trained to estimate $H$ and $V$ from temporally ordered visual features extracted from RGB video frames. The regression process involved designing transformer-based model architecture, defining data handling procedures, training and validation strategies, and evaluating model performance, each of which is described below.

#### 2.6.1 Model Architecture

A transformer-based regression model was used to estimate $H$ and $V$ distances while explicitly modeling temporal dependencies across sequences of video frames. The model architecture consisted of a bidirectional transformer encoder designed to capture both short- and long-range temporal relationships within lifting sequences. Incorporating temporal context allowed the model to leverage information from neighboring frames, which can reduce the impact of transient occlusions, intermittent ROI detection or segmentation failures, and frame-level noise, while enforcing temporally consistent distance estimates across the lifting sequence (e.g., Arnab et al., 2019). Transformer-based architectures are well suited for this task, because they use self-attention mechanisms to model dependencies across an entire sequence, rather than relying solely on local temporal context (Dosovitskiy, 2020; Vaswani et al., 2017).

Each input frame was represented by a 773-dimensional feature vector, consisting of DINOv2 visual features augmented with box-related geometric information (Section 2.5). These frame-level feature vectors were first projected into a 512-dimensional latent space using a linear embedding layer. Sinusoidal positional encodings were then added to preserve temporal ordering within each sequence. The embedded sequences were processed by a stack of six transformer encoder layers, each comprising eight attention heads, feedforward sublayers, dropout regularization, and residual connections. The transformer output at each time step was passed through a regression head consisting of fully connected layers with nonlinear activation and dropout, producing simultaneous estimations of *H* and *V* for each frame. Although transformer-based regression model produces frame-level estimates of *H* and *V* for all frames in each sequence during training, estimation performance was evaluated only for the start and end frames of each lift, consistent with RNLE parameter definitions.

### 2.6.2 Data Handling, Model Training, and Validation

Input data were organized into fixed-length sequences of 100 consecutive frames. For training, overlapping temporal windows were generated using a 50% stride (i.e., each window advanced by half its length) to increase the number of training samples and improve temporal generalization. For validation, fixed-length sequences corresponding to the start and end of each lift were extracted to align with RNLE parameter definitions. Variable-length sequences were handled through zero padding, and attention masks were applied to ensure that padded frames did not contribute to the loss. Ground-truth *H* and *V* values derived from IMU data (Section 2.3) were normalized prior to training by dividing by a fixed scalar (2000 mm), which exceeds the maximum expected range of *H* and *V* in the dataset and ensured that normalized values remained within a consistent and numerically stable range across participants. Model training was performed for up to 100 epochs using the AdamW optimizer, with a learning-rate scheduler that reduced the learning rate when validation loss plateaued (e.g., Loshchilov & Hutter, 2019; Senior et al., 2013). Early stopping was applied to prevent overfitting (Prechelt, 1998); training was terminated if validation loss failed to improve over 15 consecutive epochs. When a new minimum validation loss was achieved, the model state was saved for subsequent evaluation.

A leave-one-subject-out (LOSO) cross-validation strategy was used for model validation. Model training and validation were performed across 32 folds, with each fold using data from 31 participants for training and data from the remaining one participant held out for validation. This validation approach accounts for inter-individual variability and reflects deployment scenarios in which models trained on known individuals are applied to an unseen worker (Gholamiangonabadi et al., 2020).

### 2.6.3 Evaluation Metrics

Regression performance was evaluated separately for the start and end of a lift by comparing predicted and ground-truth *H* and *V* values. Model performance was quantified using three standard regression metrics for *H* and *V* (see Appendix A.7 for evaluation metric definitions and computation details): mean absolute error (MAE), root mean square error (RMSE), and maximum absolute error (MaxAE). For each camera-view condition and VLM-based pipeline, performance metrics were computed for each LOSO fold.

### 2.7 Statistical Analyses

To assess the effects of *Pipeline* and *Camera View Condition* on *H* and *V* estimation performance at the start and end of a lift, separate two-way repeated-measures analyses of variance (RANOVAs) were performed for each error metric. For each RANOVA model, biological sex (*Sex*) was included as a blocking effect. Where relevant, significant interaction effects were explored using simple-effects testing, and *post hoc* paired comparisons were completed using the Tukey's HSD procedure. All statistical analyses were performed with JMP Pro 18 (SAS, Cary, NC) using the restricted maximum likelihood (REML) method. To obtain normal distributions of model residuals, logarithmic transformations were applied to six dependent variables—MAE of *H* at the start and end of a lift, MAE and RMSE of *V* at the end of a lift, and MaxAE of *H* and *V* at the end of a lift. Statistical significance was determined when $p < 0.05$, and summary data are reported as least-square means based on the statistical model fits.

## 3.0 Results

All RANOVA results are summarized in Tables A.1 and A.2. Significant main or interaction effects of *Pipeline* and *Camera View Condition* were found across all regression performance metrics. More detailed results are provided below.

### 3.1 Start of a Lift: Estimation Errors of Horizontal and Vertical Distances

Significant main effects of *Pipeline* and *Camera View Condition* were found across all error metrics (i.e., MAE, RMSE, and MaxAE) for both *H* and *V* estimation (Table A.1). Additionally, significant *Pipeline* × *Camera View Condition* interaction effects were found for MAE of both *H* and *V* estimations, RMSE of *V* estimation, and MaxAE of *V* estimation. Across all metrics, the GD–SAM–Dv2 pipeline yielded significantly smaller errors vs. the GD–Dv2 pipeline. When estimating *H*, mean values were MAE = ~7.2 cm vs. ~9.25 cm; RMSE = ~9.6 cm vs. ~12.1 cm; and MaxAE = ~20.7 cm vs. ~23.6 cm. When estimating *V*, mean values were MAE = ~14.5 cm vs. ~23.0 cm; RMSE = ~18.3 cm vs. ~27.7 cm; and MaxAE = ~40.7 cm vs. ~49.0 cm.

Across all metrics, the V1+V2+V3 multi-view condition resulted in significantly smaller errors vs. single-view condition for both *H* and *V* estimation. When estimating *H*, V1+V2+V3 produced the smallest errors: MAE = ~6.2 cm; RMSE = ~8.1 cm; MaxAE = ~19.8 cm, while single-view conditions led to the largest errors, with V1 producing: MAE = ~10.68 cm; RMSE = ~13.5 cm; MaxAE = ~24.6 cm. When estimating *V*, V1+V2+V3 similarly generated the smallest errors: MAE = ~7.78 cm; RMSE = ~11.0 cm; MaxAE = ~37.2 cm, while single-view conditions yielded the greatest errors, with V2 producing MAE = ~27.73 cm and RMSE = ~32.6 cm, and V1 producing MaxAE = ~51.6 cm.

Generally, the magnitude of the difference in estimation error between the two pipelines varied, and in some cases significantly, across camera view configurations (Figures 4, A.4, and A.5). When estimating *H*, the GD–SAM–Dv2 pipeline with multi-view conditions consistently produced the smallest errors: MAE = ~6.0 cm for V1+V3, V2+V3, and V1+V2+V3 (Figure 4); RMSE = ~7.9–8.0 cm for V2+V3, V1+V3, V1+V2, and V1+V2+V3; and MaxAE = ~18.8–19.4 cm for V2+V3, V1+V2+V3, V1+V2, and V1+V3. In contrast, the GD–Dv2 pipeline with the V1 single-view condition yielded the largest errors, with MAE = ~12.39 cm (Figure 4), RMSE = ~15.1 cm, and MaxAE = ~26.4 cm.

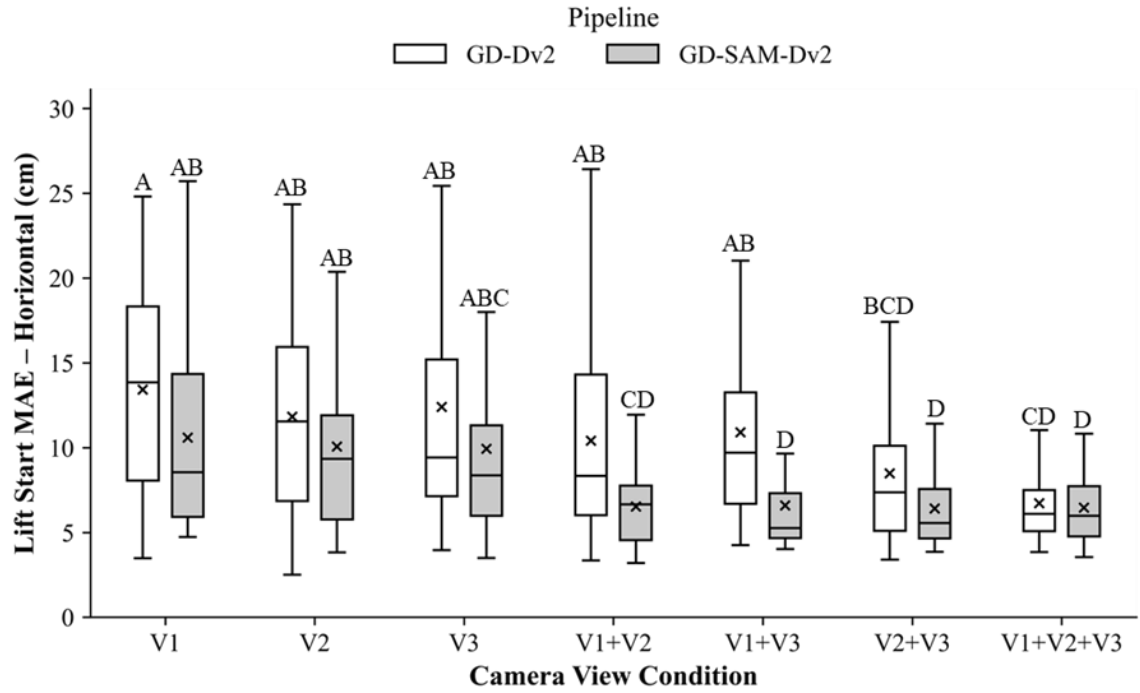

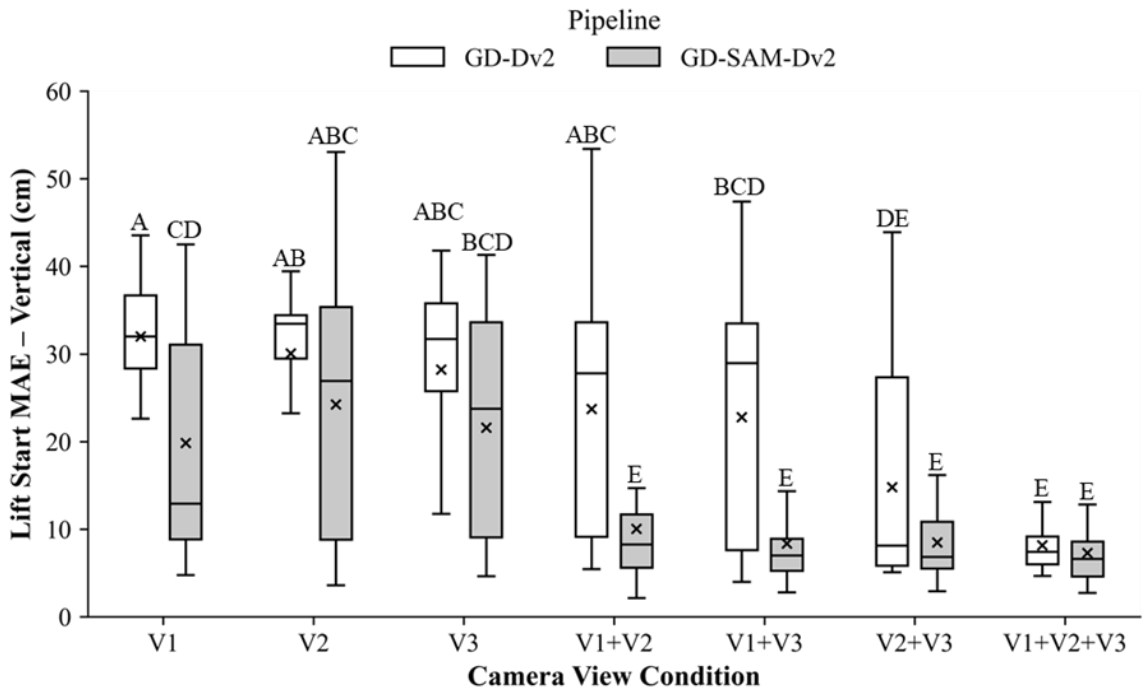

Figure 4: Significant interaction effect of *Pipeline × Camera View Condition* on MAE of *H* (top) and *V* (bottom) estimation at the start of a lift. Hereafter, for each condition, mean estimation errors were computed for each participant across all trials within each fold, and the box plots summarize the distribution of these participant-level mean errors. Based on *post hoc* paired comparisons, conditions that do not share a common letter are significantly different; this lettering convention is used for all subsequent figures.

When estimating *V*, the GD–SAM–Dv2 pipeline with multi-view conditions yielded the smallest errors, with MAE = ~7.35 cm for V1+V2+V3 (Figure 4), RMSE = ~10.7 cm for V1+V2+V3 (Figure A.4), and MaxAE = ~35.6 cm for V1+V3 (Figure A.5). In contrast, the GD–Dv2 pipeline with the V1 single-view condition produced the largest errors, with MAE = ~32.39 cm (Figure 4), RMSE = ~38.3 cm (Figure A.4), and MaxAE = ~58.0 cm (Figure A.5).

The absolute difference between the smallest- and largest-error pipeline–camera combinations was substantially larger for *V* estimation vs. for *H* estimation, with MAE = ~25.0 cm vs. ~6.4 cm (Figure 4), RMSE = ~27.6 cm vs. ~7.2 cm, and MaxAE = ~22.4 cm vs. ~7.6 cm, indicating greater variability in *V* estimation performance.

Across camera view conditions, the magnitude of error reduction associated with the GD–SAM–Dv2 pipeline was generally larger for single-view conditions vs. multi-view configurations. When estimating *H*, error reductions were larger for single-view vs. multi-view conditions, with MAE = ~2–3 cm vs. ~1–2 cm, RMSE = ~2–4 cm vs. ~1–3 cm, and MaxAE = ~3–5 cm vs. ~2–3 cm. When estimating *V*, error reductions were likewise larger for single-view vs. multi-view conditions, with MAE = ~8–15 cm vs. ~4–8 cm, RMSE = ~9–16 cm vs. ~5–10 cm, and MaxAE = ~8–13 cm vs. ~4–10 cm.

### 3.2 End of a Lift: Estimation Errors of Horizontal and Vertical Distances

Significant main effects of *Pipeline* and *Camera View Condition*, as well as significant *Pipeline × Camera View Condition* interaction effects, were found across all error metrics for both *H* and *V* estimation, except for the main effect of *Camera View Condition* and the *Pipeline × Camera View Condition* interaction effect on MaxAE for *H* estimation (Table A.2). Across all metrics, the GD–SAM–Dv2 pipeline yielded significantly smaller errors vs. the GD–Dv2 pipeline. When estimating *H*, mean values were MAE = ~9.3 cm vs. ~13.2 cm, RMSE ~13.1 cm vs. ~17.0 cm, and MaxAE ~27.0 cm vs. ~30.5 cm. When estimating *V*, mean errors were MAE = ~7.5 cm vs. ~12.2 cm, RMSE = ~8.8 cm vs. ~13.5 cm, and MaxAE = ~18.0 cm vs. ~22.3 cm.

Where significant main effects of *Camera View Condition* were found, the V1+V2+V3 multi-view configuration generally yielded smaller errors vs. single-view configurations for both *H* and *V* estimation. When estimating *H*, V1+V2+V3 resulted in the smallest errors, with MAE = ~8.1 cm and RMSE = ~10.9 cm, whereas single-view conditions produced larger errors, with the largest MAE and RMSE occurring for the V2 condition, in which MAE = ~14.4 cm and RMSE = ~18.5 cm. When estimating *V*, V1+V2+V3 produced the smallest errors across all metrics, with MAE = ~4.8 cm, RMSE = ~6.0 cm, and MaxAE = ~15.0 cm, whereas single-view conditions produced the largest errors, especially for the V1 condition, with MAE = ~15.5 cm, RMSE = ~16.7 cm, and MaxAE = ~24.9 cm.

Generally, the magnitude of the difference in estimation error between the two pipelines varied, and in some cases significantly, across camera view configurations. When estimating *H*, the GD–SAM–Dv2 pipeline with multi-view configurations generally produced smaller errors, with MAE = ~7.4–7.6 cm for the V1+V2+V3 and V1+V2 conditions (Figure 5) and RMSE = ~10.6–10.9 cm for the V1+V2+V3, V1+V3, V2+V3, and V1+V2 conditions (Figure A.6). In contrast, the GD–Dv2 pipeline with single-view conditions produced the largest errors, most notably for the V1 condition, with MAE = ~16.7 cm (Figure 5) and RMSE = ~20.2 cm (Figure A.6).

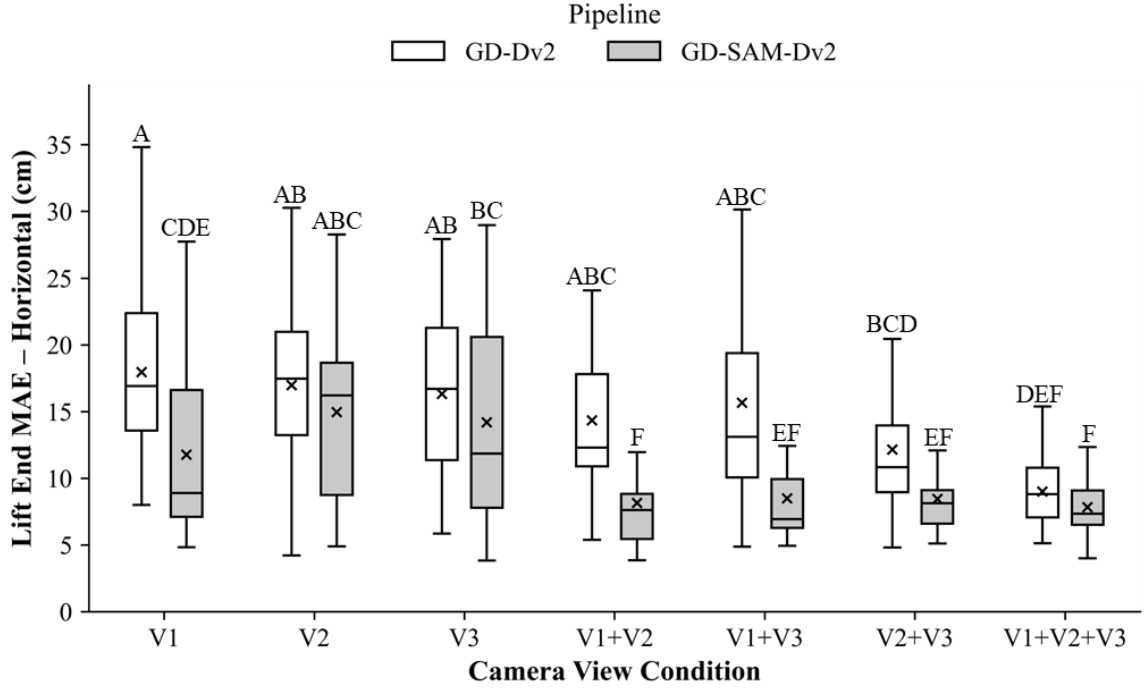

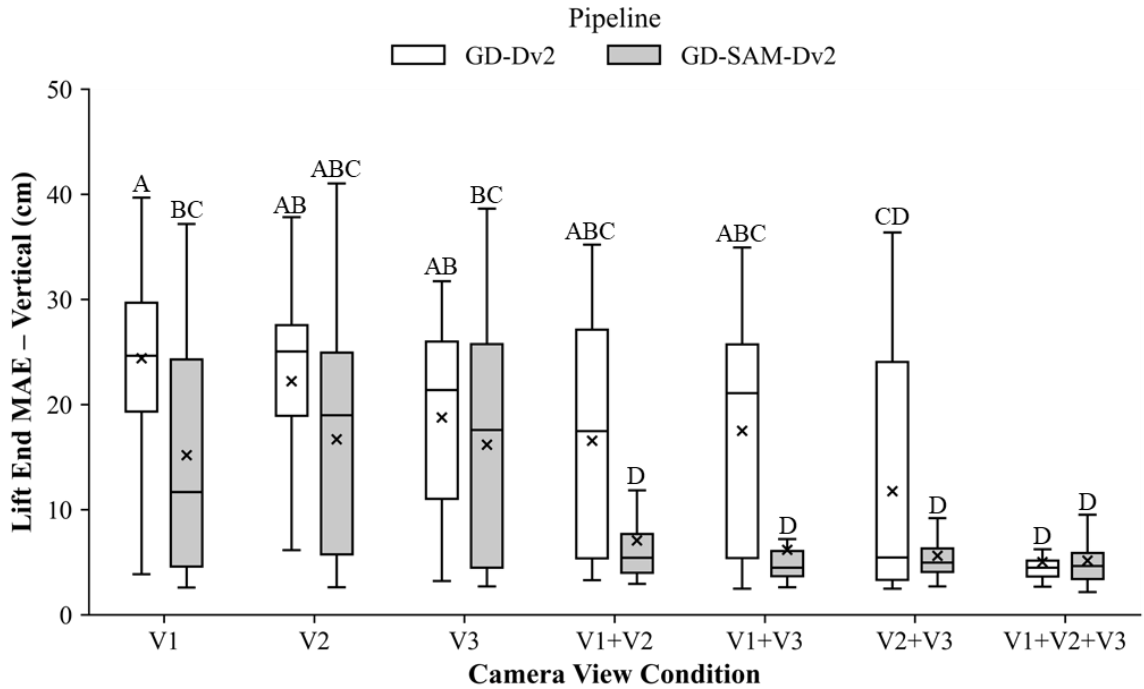

Figure 5: Significant interaction effect of *Pipeline* × *Camera View Condition* on MAE of *H* (top) and *V* (bottom) estimation at the end of a lift.

When estimating *V*, the GD–SAM–Dv2 pipeline with multi-view configurations generally produced the smallest errors, with MAE = ~4.9–5.9 cm across all multi-view conditions (Figure 5), RMSE = ~6.0–7.2 cm across all multi-view conditions (Figure A.7), and MaxAE = ~14.9–15.1 cm for the V1+V2+V3 and V1+V3 conditions (Figure A.8). In contrast, the GD–Dv2 pipeline with single-view conditions produced the largest errors, most notably for the V1 condition, with MAE = ~22.7 cm (Figure 5), RMSE = ~23.1 cm (Figure A.7), and MaxAE = ~29.4 cm (Figure A.8).

The absolute difference between the smallest- and largest-error pipeline–camera combinations was larger for *V* estimation vs. *H* estimation, with MAE = ~17.8 cm vs. ~9.3 cm (Figure 5), RMSE = ~17.1 cm vs. ~9.6 cm (Figures A.6 and A.7), and MaxAE = ~14.6 cm vs. ~3.0 cm, indicating greater variability in *V* estimation performance.

Across camera configurations, the magnitude of error reduction for the GD–SAM–Dv2 pipeline was generally larger for single-view conditions vs. multi-view configurations. When estimating *H*, larger error reductions were found for single-view vs. multi-view conditions, with MAE = ~4–6 cm vs. ~1–2 cm, RMSE = ~5–9 cm vs. ~2–4 cm, and MaxAE = ~1–5 cm vs. ~1–4 cm. When estimating *V*, larger error reductions were likewise found for single-view vs. multi-view conditions, with MAE = ~8–13 cm vs. ~2–5 cm, RMSE = ~7–11 cm vs. ~2–6 cm, and MaxAE = ~5–9 cm vs. ~1–5 cm.

## 4.0 Discussion

Our goal here was to evaluate the feasibility of using VLM-based pipelines to estimate *H* and *V* required by the RNLE from RGB video. Our findings provide new results that demonstrating that VLM-based pipelines combining text-guided object detection, pixel-level segmentation, and temporal modeling can provide effective and ambient estimates of RNLE distance parameters. Estimation performance varied across VLM-based pipelines and camera view conditions, though, with particularly notable differences between the start and end phases of lifting tasks. Below, we discuss key findings in relation to pipeline design, camera view condition, temporal dynamics, and practical applications for ergonomic assessment.

### 4.1 Effect of Pixel-Level Segmentation on *H* and *V* Estimation Performance

The GD–SAM–Dv2 pipeline, which used pixel-level segmentation, generally yielded smaller errors vs. the GD–Dv2 pipeline across most error metrics for both *H* and *V* estimation at the start and end of a lift. Overall, segmentation produced consistent error reductions of ~20–30% when estimating *H* and ~35–40% when estimating *V* (Figures 4 and 5). These results are consistent with the notion that pixel-level segmentation provides more precise spatial representations of task-relevant regions than does bounding box–based localization alone (Kirillov et al., 2023; Ren et al., 2024). Bounding boxes inherently include background pixels and may fail to accurately capture the spatial extent of irregularly shaped objects or body segments, particularly when the lifter's posture varies substantially across frames or when the handled load partially occludes body segments. By contrast, SAM-generated segmentation masks isolate object boundaries at the pixel level (Kirillov et al., 2023), enabling the feature extractor (i.e., DINOv2) to focus on semantically relevant visual content while suppressing background interference. Recent work has demonstrated similar benefits of integrating SAM with object detection models for fine-grained spatial localization tasks (e.g., Han et al., 2023; Ren et al., 2024; Wang et al., 2024), suggesting that pixel-level segmentation can improve feature quality for downstream classification or regression tasks.

Consistent with those earlier findings, the GD–SAM–Dv2 pipeline here yielded the most substantial segmentation-related improvements for *V* estimation relative to the GD–Dv2 pipeline, with errors ~8.5 cm smaller at the start and ~4.7 cm smaller at the end of a lift. The greater benefit for *V* estimation may reflect the sensitivity of vertical distance measures to precise localization of the hands and feet, consistent with principles of perspective projection, in which small vertical displacements in image space can correspond to substantial differences in physical height when projected into real-world coordinates (Hartley & Zisserman, 2003). The advantage of pixel-level segmentation was most evident under single-view camera conditions, in which the GD–SAM–Dv2 pipeline yielded substantially larger error reductions vs. multi-view conditions. For instance, at the start of a lift, the addition of segmentation yielded *V* estimation MAE values ~8–15 cm smaller for single-view conditions, but ~4–8 cm smaller for multi-view conditions. This pattern suggests that segmentation improves object localization within individual frames, which is particularly beneficial when viewpoint diversity is limited (Bae et al., 2022; Pavllo et al., 2018). In comparison, under multi-view conditions the added benefit of segmentation is smaller, but still meaningful, with MAE reductions of ~1–2 cm when estimating *H* and ~4–8 cm when estimating *V*.

**4.2 Effect of Camera View Condition on *H* and *V* Estimation Performance**

Camera view conditions varied with *H* and *V* estimation performance, with multi-view conditions generally yielding smaller errors vs. single-view conditions. At the start of a lift, the V1+V2+V3 condition produced MAE values of ~6.2 cm when estimating *H* and ~7.8 cm when estimating *V*, which were significantly smaller vs. the worst-performing single-view conditions, with MAE = ~10.7 cm for the V1 condition when estimating *H* and MAE = ~27.7 cm for the V2 condition when estimating *V* (Figure 4). This performance advantage of the V1+V2+V3 condition was maintained at the end of a lift, with MAE = ~8.1 cm when estimating *H* and MAE = ~4.8 cm when estimating *V*, which were significantly smaller vs. the worst-performing single-view conditions, with MAE = ~14.4 cm for the V2 condition when estimating *H* and MAE = ~15.5 cm for the V1 condition when estimating *V* (Figure 5). These results suggest that multi-view capture provides geometric redundancy that reduces localization uncertainty caused by occlusion and perspective effects in monocular video (Hartley & Zisserman, 2003; Remondino & El-Hakim, 2006; Szeliski, 2022). Similar benefits of using multiple synchronized views have been reported in multi-view human pose estimation, in which incorporating information from more than one camera yields more accurate spatial localization of body landmarks than single-view approaches, particularly in conditions involving occlusion or unfavorable viewing angles (Qiu et al., 2019). This pattern is consistent with prior evidence that camera placement meaningfully affects vision-based ergonomic assessment outputs, with viewpoint-dependent visibility and depth ambiguity contributing to variability in estimated RNLE-related quantities (Murugan et al., 2025; Neupane et al., 2024; Wang et al., 2019; Zahabi et al., 2026; Zhang et al., 2022).

In the present dataset, V1 and V2 corresponded to two oblique views of the workspace from opposite sides, whereas V3 provided a frontal view of the participant. When synchronized views with differing perspectives were used, multi-view conditions showed more consistent localization of task-relevant landmarks used to estimate RNLE distances, including the hands, shoes, and handled load, across frames. This redundancy was particularly relevant for *V* estimation, which exhibited greater variability under single-view conditions and showed larger error reductions under multi-view conditions. Notably, prior validation work indicates that RNLE distance-related inputs are less accurate under single-view capture and benefit from frontal or sagittal camera perspectives, supporting the inclusion of a frontal view (V3) in multi-view conditions (Neupane et al., 2024; Wang et al., 2019; Zahabi et al., 2026; Zhang et al., 2022).

Interestingly, certain two-camera conditions, particularly V1+V3 and V2+V3, produced errors comparable to the V1+V2+V3 condition when estimating *H* at the start of a lift, most notably for MAE and RMSE. When estimating *V*, these two-camera conditions likewise showed comparable performance to V1+V2+V3 for MAE at the start of a lift, although differences became more pronounced at the end of a lift. Notably, both V1+V3 and V2+V3 combine a frontal view with an oblique view, suggesting that these pairing can capture much of the task-relevant geometric information required for RNLE distance estimation, particularly during the initial phase of a lift. In these conditions, the frontal view supports localization of vertical displacement, while the oblique view preserves information related to anterior–posterior and mediolateral positioning. However, the V1+V2+V3 condition consistently yielded the smallest errors across metrics and lift phases, indicating more stable performance even when differences in mean error were relatively small. These findings suggest that multi-view camera setups can provide more stable *H* and *V* estimation, particularly for *V*, although the magnitude of improvement depends on the specific combination of views. This stability matters, because errors in these inputs can propagate to risk outputs; for example, there is recent evidence of systematic bias in RNLE metrics derived using a commercial tool (e.g., RWL overestimation and LI underestimation on average), implying that reducing view-dependent uncertainty in *H* and *V* can affect risk classification rather than merely improving kinematic fidelity (Zahabi et al., 2026).

**4.3 Temporal Differences in *H* and *V* Estimation Performance Between Start and End of Lift**

Estimation errors for *H* and *V* exhibited opposite temporal patterns between the start and end of a lift. Specifically for the GD–SAM–Dv2 pipeline, *V* estimation errors decreased substantially from the start to the end of a lift, with respective MAE of ~14.5 vs. ~7.5 cm, whereas *H* errors increased, with respective values of ~7.2 vs. ~9.3 cm (see Figure A.9 for MAE comparison between start and end of a lift). This divergence persisted across both pipelines and most camera view conditions, indicating that the underlying error sources for *H* and *V* might vary across the lifting phases. At the start of a lift, participants adopted a posture involving torso flexion, with their hands near the floor, and the hands and lower body segments were more likely to be partially occluded by the torso and handled load, particularly in the oblique single-view conditions (i.e., V1 and V2) and in multi-view conditions that relied primarily on oblique perspectives (i.e., V1+V2). These occlusions can reduce the consistency of ROI localization and degrade frame-level estimation, particularly for *V*.

At the end of a lift, participants were more upright with their hands near hip height, which might reduce occlusion and improve visibility of the hands, especially when a frontal perspective was available (i.e., V3, V1+V3, V2+V3, and V1+V2+V3), supporting smaller *V* estimation errors. In contrast, the increase in *H* estimation error from the start to the end of a lift may reflect greater difficulty localizing ankle-related landmarks at the end of a lift. As the load is raised, it may partially occlude the lower body or reduce the visibility of shoe-related ROIs used to localize the ankles, which would disproportionately affect *H* estimation. This ankle-visibility limitation is most relevant for single-view conditions—particularly V1 and V2—and was mitigated, but not eliminated, when synchronized views include the frontal perspective (i.e., V1+V3, V2+V3, and V1+V2+V3). Overall, these findings indicate that *H* and *V* estimation performance is not uniform across the start and end of a lift and is sensitive to posture-dependent occlusion patterns. This pattern suggests that a single regression model may not optimally capture both lift phases.

### 4.4 Practical Applications for VLM-Based Ergonomic Assessment

The present findings have several implications for future applications of VLM-based systems in occupational ergonomic assessment. First, the demonstrated feasibility of estimating RNLE distance parameters from RGB video—without requiring manual measurement, wearable sensors, or marker-based motion capture systems—supports the use of VLM-based approaches to enable video-based ergonomic risk assessments. Second, multi-view camera conditions consistently yielded smaller estimation errors than single-view conditions, indicating that practical implementations should prioritize the use of multiple synchronized views when feasible. While single-view conditions reduced hardware and installation cost, they resulted in substantially larger estimation errors—particularly for $V$—which may limit their suitability for applications requiring reliable distance estimates. Also, two-camera conditions (e.g., V1+V3) yielded estimation performance comparable to three-camera configurations in several cases, indicating a favorable trade-off between estimation performance and system complexity that may be appropriate in resource-constrained environments.

Third, the consistent advantage of pixel-level segmentation across camera view conditions suggests that segmentation-augmented pipelines should be prioritized in future implementations when estimation performance is a primary concern. Although the detection-only pipeline may reduce computational complexity, the segmentation-based approach yielded substantial reductions in estimation error (approximately 22–39% across metrics), which may justify the additional computational cost in many ergonomic assessment applications. As segmentation models continue to improve in efficiency (e.g., Ravi et al., 2024), this trade-off is likely to become increasingly favorable.

### 4.5 Limitations and Future Research Directions

Several limitations of our study should be acknowledged. Although the estimation performance was evaluated across multiple participants of both biological sexes with varied anthropometries and across a range of lifting conditions (e.g., lift origin, hand configuration, and box mass), all tasks were performed by relatively young (18–39 years old), healthy participants in controlled laboratory settings with uniform backgrounds and consistent lighting. Thus, caution should be taken in generalizing the findings to other populations and to real-world occupational settings. In practice, work environments often involve variable lighting, cluttered backgrounds, dynamic occlusions from coworkers or equipment, and more diverse lifting postures, all of which may affect model performance. Although we relied on zero-shot VLMs for detection and segmentation of ROIs and feature extraction (i.e., Grounding DINO, SAM, and DINOv2) without task-specific fine-tuning—suggesting some degree of generalizability—future work should explicitly evaluate model performance in real environments to better assess external validity. In addition, ground-truth distance labels were derived from wearable IMU sensors, which introduce their own measurement uncertainty and may not perfectly align with the anatomical landmarks specified by the RNLE. While IMU-based motion capture has been validated for kinematic measurements (Schall et al., 2018), small errors related to sensor placement relative to true joint centers may propagate into the reference labels. Future studies could incorporate marker-based optical motion capture or direct physical measurements to provide independent validation of VLM-based estimates.

Further, while our results demonstrate the feasibility of estimating *H* and *V*, the RNLE includes additional parameters—such as asymmetry angle, coupling quality, and task frequency—that were not addressed here. Future work could examine whether VLM-based pipelines can be extended to estimate these additional RNLE parameters directly from video, enabling ambient RNLE-based risk assessment. Finally, differences in estimation performance between the start and end phases of lifts suggest that a single regression model may not optimally capture the full range of lifting postures. Future work could explore phase-aware modeling strategies or architectures that explicitly account for posture changes across the lifting tasks.

## 5.0 Conclusions

Efficiently and accurately estimating RNLE distance parameters (i.e., *H* and *V*) during manual lifting is essential for assessing and controlling the risk of WMSDs. While a range of approaches have been used for ergonomic assessment, including observation-based methods, wearable sensors, marker-based motion capture systems, and pose-based vision methods, these approaches are often time-consuming, intrusive, or difficult to scale in real-world work environments. We developed and evaluated VLM-based pipelines to non-invasively estimate *H* and *V* from RGB video streams, including a text-guided, detection-only pipeline and a detection-plus-segmentation pipeline, each of which was assessed across seven camera view conditions using leave-one-subject-out validation. Estimation performance varied across pipelines and camera view conditions, with the segmentation-based, multi-view pipeline consistently yielding the lowest errors, achieving mean absolute errors of approximately 6–8 cm when estimating *H* and 5–8 cm when estimating *V*. Overall, these findings demonstrate the feasibility of using VLM-based pipelines to provide video-based estimates of RNLE distance parameters, supporting progress toward ergonomic risk assessment without wearable sensors or manual measurements. However, further research is needed to evaluate pipeline performance across diverse worker populations and in real occupational settings.

## 6.0 Declaration of generative AI and AI-assisted technologies in the manuscript preparation process

During the preparation of this work, we used ChatGPT to refine some sentences, improve the clarity of the text, and generate a schematic illustration (i.e., Figure 2). After using this tool/service, we reviewed and edited the content as needed, and we take full responsibility for the content of this publication.

## 7.0 Acknowledgements

The first author was supported by a predoctoral training program grant (T03 OH008613) from CDC/NIOSH. The current contents are solely the authors' responsibility and do not necessarily represent the official views of CPWR, NIOSH, or the CDC. The authors thank Advanced Research Computing at Virginia Tech for providing computational resources and technical support that contributed to the results reported in this paper. URL: https://arc.vt.edu/

# Appendix

## A.1 Participants

A convenience sample of 32 young adults (19 males and 13 females) completed the study and was recruited from the university and local community. Respective means (SD) of age, body mass, and height were 26.8 (4.5) years, 77.1 (12.2) kg, and 176.0 (5.7) cm for the males; and 27.0 (5.6) years, 68.5 (8.4) kg, and 169.3 (6.7) cm for the females. All participants self-reported being right-handed, physically active (i.e., exercising at least twice per week), and having no musculoskeletal disorders within the past year. The research reported herein compiled with the tenets of the Declaration of Helsinki, and the study protocol was approved by the Institutional Review Board at Virginia Tech (#23-095). Informed consent was obtained from all participants prior to any data collection (Ojelade et al., 2025; Ojelade, 2024).

## A.2 Symmetric Lifting Task Details

A single wood box (width = 26.0 cm; depth = 41.0 cm; and height = 23.5 cm; Figure A.1) was used for all symmetric lifting trials.

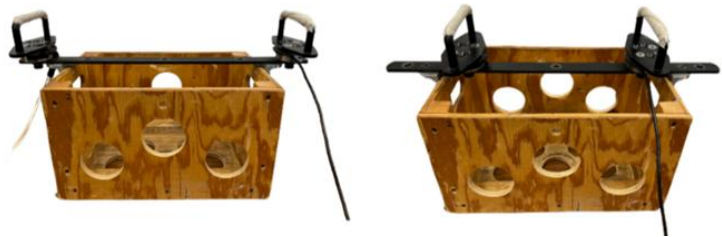

Figure A.1. Illustration of the two hand configurations: broad (left), and narrow (right; adapted from Ojelade et al. (2025)

## A.3 Experimental Procedures

A repeated-measures experimental design was implemented. Participants completed a training phase to practice the tasks using their comfortable work strategies and speed, simulating an industrial setting. During the experimental phase, the order of *Hand Configurations* and *Box Mass* presentation was counterbalanced using balanced Latin square designs to reduce potential bias. To minimize physical fatigue, a mandatory rest period of at least four minutes was provided between trials involving different hand configurations (Ojelade et al., 2025).

## A.4 Azure Kinect™ Camera Instrumentation

Whole-body motion was recorded using three synchronized Azure Kinect™ markerless camera systems (Microsoft Corporation, Seattle, WA, USA), sampled at 30 Hz. The cameras were positioned approximately 1.74 m from the edge of the work area; this configuration was established during pilot testing to improve coverage given the narrow field of view of the camera units. The three cameras were time-synchronized using a 3.5-mm auxiliary cable connected in a daisy-chain configuration, with one camera designated as the primary device and the other two as secondary devices (Ojelade et al., 2025).

**A.5 Wearable IMU Data Processing**

Whole-body kinematics were captured from a wearable inertial motion capture system (Noraxon Ultium, Noraxon, Scottsdale, AZ, USA) and were exported using Noraxon myoResearch. To minimize temporal offsets among motion capture systems, the IMU system was synchronized with other data sources using an analog output signal generated from LabView. Whole-body kinematics were exported in Biovision Hierarchy (BVH) format from Noraxon myoResearch. Exported kinematics were low-pass filtered (6 Hz cutoff; 4th-order Butterworth; bidirectional), and kinematics data were downsampled to 30 Hz to match the sampling rate of the Azure Kinect™ cameras (Ojelade et al., 2025).

**A.6 Revised NIOSH Lifting Equation Definition**

The Revised NIOSH Lifting Equation (RNLE) is a biomechanical risk assessment model to evaluate the physical demands of two-handed manual lifting tasks and to estimate the associated risk of low-back musculoskeletal injury (Waters et al., 1993). The RNLE provides a recommended weight limit (RWL) for a given lifting task based on task geometry, load characteristics, and temporal factors, and it forms the basis for computing the Lifting Index (LI), a commonly used indicator of lifting risk. The RNLE expresses the recommended weight limit as:

$$RWL = LC \times HM \times VM \times DM \times AM \times FM \times CM$$

where $LC$ is the load constant and the remaining multipliers account for task-specific characteristics, including horizontal location of the hands (*HM*), vertical location of the hands (*VM*), vertical travel distance (*DM*), asymmetry (*AM*), lift frequency (*FM*), and coupling quality (*CM*). Among these factors, horizontal distance (*H*) and vertical distance (*V*) are fundamental geometric inputs that directly affect several multipliers in the equation and estimated biomechanical loading. In the RNLE, *H* and *V* describe the spatial relationship between the worker's hands, body, and the floor at the start and end of each lift (Fox et al., 2019; Waters et al., 1993).

**A.7 Evaluation Metrics Details**

*H* and *V* estimation performance was evaluated separately for the start and end of the lift by comparing predicted ($\hat{y}_i$) and ground-truth ($y_i$) values. For each lift trial, error metrics were computed across all start and end of the lift frames (indexed by $i = 1,…, N$) for each participant fold, camera-view condition, and VLM-based pipeline.

$$Mean\ Absolute\ Error\ (MAE) = \frac{1}{N}\sum_{i=1}^{N} |\hat{y}_i - y_i|$$

$$Root\ Mean\ Square\ Error\ (RMSE) = \sqrt{\frac{1}{N}\sum_{i=1}^{N}(\hat{y}_i - y_i)^2}$$

$$Maximum\ Absolute\ Error\ (MaxAE) = \max_{i \in \{1,…,N\}} |\hat{y}_i - y_i|$$

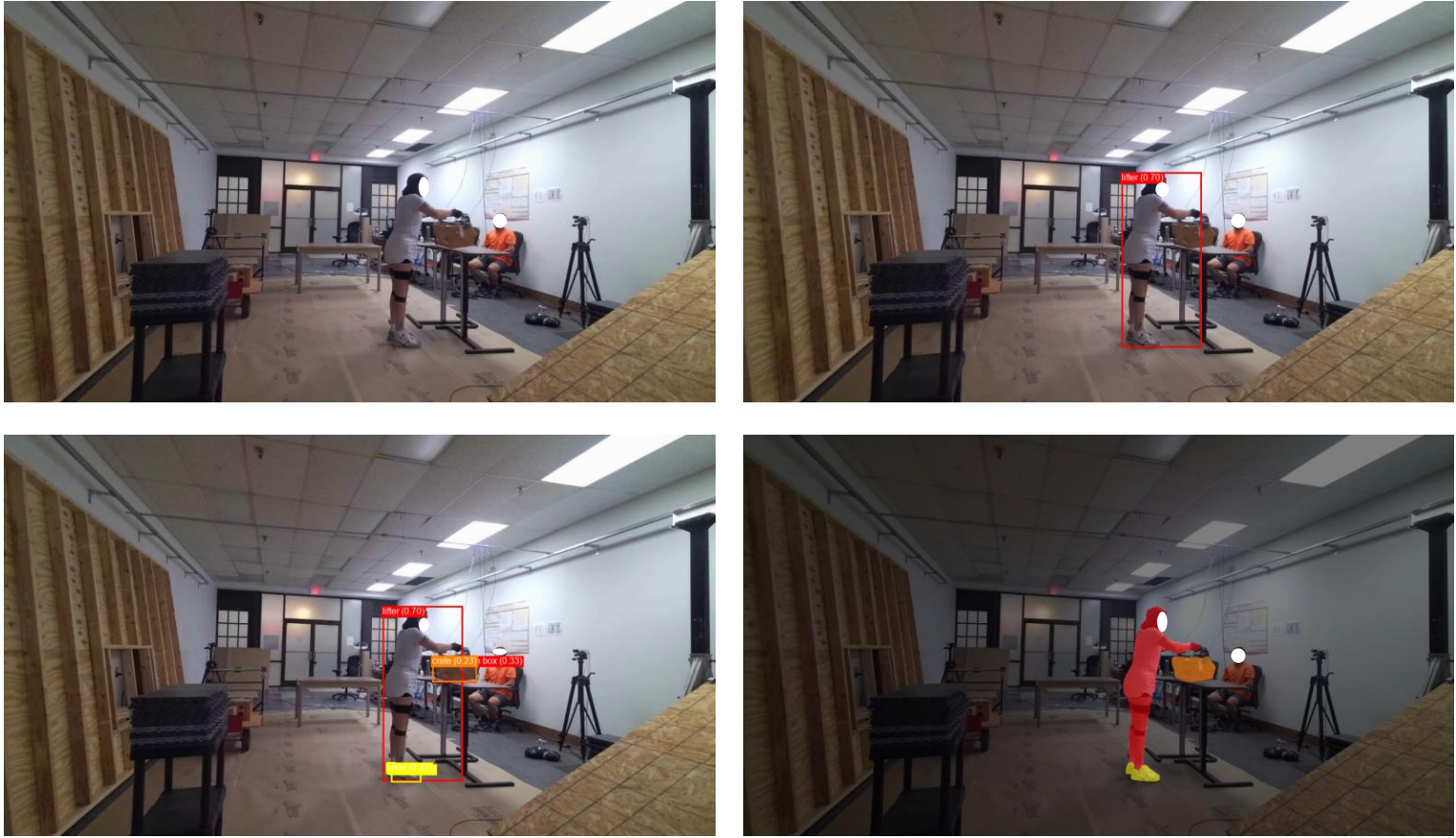

Figure A.2: Example visualization of the ROI detection and segmentation process for a representative lifting frame of Camera View 2 (V2). Top left: Original RGB video frame. Top right: Detection of the primary lifter using Grounding DINO, with the bounding box corresponding to the highest-confidence person detection. Bottom left: Detection of task-relevant ROIs within a cropped region centered on the lifter using Grounding DINO. Bottom right: Pixel-level segmentation of the detected ROIs produced by the Grounding DINO + SAM pipeline.

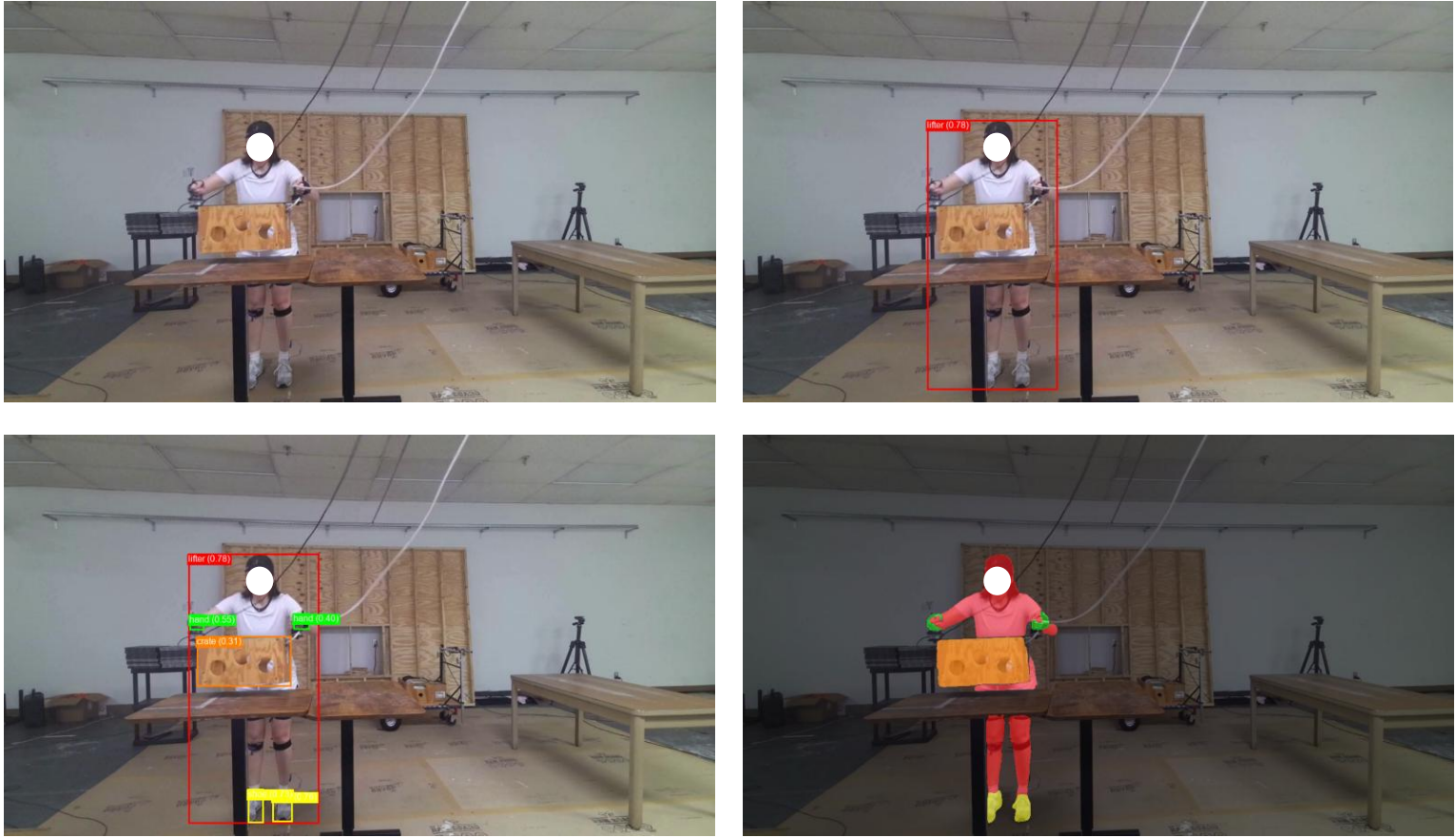

Figure A.3: Example visualization of the ROI detection and segmentation process for a representative lifting frame of Camera View 3 (V3). Top left: Original RGB video frame. Top right: Detection of the primary lifter using Grounding DINO, with the bounding box corresponding to the highest-confidence person detection. Bottom left: Detection of task-relevant ROIs within a cropped region centered on the lifter using Grounding DINO. Bottom right: Pixel-level segmentation of the detected ROIs produced by the Grounding DINO + SAM pipeline.

Table A.1. ANOVA results assessing the effect of *Pipeline*, *Camera View Condition*, and *Sex* on estimation errors for horizontal and vertical distances at the **start of the lift**. Entries are $F$ values ($p$ values), and significant effects are highlighted in bold font.

| Effect | MAE – Horizontal | MAE – Vertical | RMSE – Horizontal | RMSE – Vertical | MaxAE – Horizontal | MaxAE – Vertical |
|---|---|---|---|---|---|---|
| *Pipeline (P)* | **40.21 (<.0001)** | **81.94 (<.0001)** | **39.19 (<.0001)** | **79.21 (<.0001)** | **25.37 (<.0001)** | **34.08 (<.0001)** |
| *Camera View Condition (C)* | **15.38 (<.0001)** | **38.05 (<.0001)** | **14.89 (<.0001)** | **35.39 (<.0001)** | **5.84 (<.0001)** | **8.77 (<.0001)** |
| *Sex (S)* | 0.10 (0.7566) | 2.62 (0.1161) | 0.10 (0.7563) | 1.90 (0.1786) | 0.00 (0.9911) | 0.00 (0.9495) |
| S × P | 0.54 (0.4646) | 0.12 (0.7253) | 0.34 (0.5587) | 0.20 (0.6518) | 0.58 (0.4485) | 1.43 (0.2319) |
| S × C | 1.10 (0.3600) | 0.68 (0.6690) | 1.08 (0.3727) | 0.69 (0.6547) | 1.21 (0.2993) | 0.96 (0.4507) |
| P × C | **2.43 (0.0257)** | **4.62 (0.0002)** | 1.89 (0.0821) | **4.65 (0.0001)** | 1.30 (0.2560) | **2.85 (0.0100)** |
| S × P × C | 1.72 (0.1146) | 1.99 (0.0664) | 1.69 (0.1231) | **2.14 (0.0483)** | 1.60 (0.1464) | 1.93 (0.0744) |

Table A.2. ANOVA results assessing the effect of *Pipeline*, *Camera View Condition*, and *Sex* on estimation errors for horizontal and vertical distances at the **end of the lift**. Entries are $F$ values ($p$ values), and significant effects are highlighted in bold font.

| Effect | MAE – Horizontal | MAE – Vertical | RMSE – Horizontal | RMSE – Vertical | MaxAE – Horizontal | MaxAE – Vertical |
|---|---|---|---|---|---|---|
| *Pipeline (P)* | **100.79** **(<.0001)** | **59.45** **(<.0001)** | **65.56** **(<.0001)** | **58.69** **(<.0001)** | **25.18** **(<.0001)** | **27.47** **(<.0001)** |
| *Camera View Condition (C)* | **20.94** **(<.0001)** | **29.92** **(<.0001)** | **18.24** **(<.0001)** | **27.84** **(<.0001)** | 1.45 (0.1952) | **11.84** **(<.0001)** |
| *Sex (S)* | 0.29 (0.5968) | 2.20 (0.1483) | 0.03 (0.8734) | 1.83 (0.1857) | 0.12 (0.7346) | 0.29 (0.5973) |
| S × P | 0.45 (0.5049) | 1.45 (0.2288) | 0.32 (0.5722) | 1.98 (0.1605) | 0.90 (0.3437) | **5.15** **(0.0238)** |
| S × C | 0.96 (0.4493) | 0.70 (0.6528) | 0.71 (0.6410) | 0.63 (0.7039) | 0.67 (0.6764) | 0.76 (0.6012) |
| P × C | **4.83** **(<.0001)** | **4.21** **(0.0004)** | **4.13** **(0.0005)** | **4.14** **(0.0005)** | 1.74 (0.1104) | **2.42** **(0.0260)** |
| S × P × C | 1.86 (0.0858) | 2.09 (0.0531) | 1.78 (0.1024) | **2.34** **(0.0309)** | 1.31 (0.2534) | **2.73** **(0.0131)** |

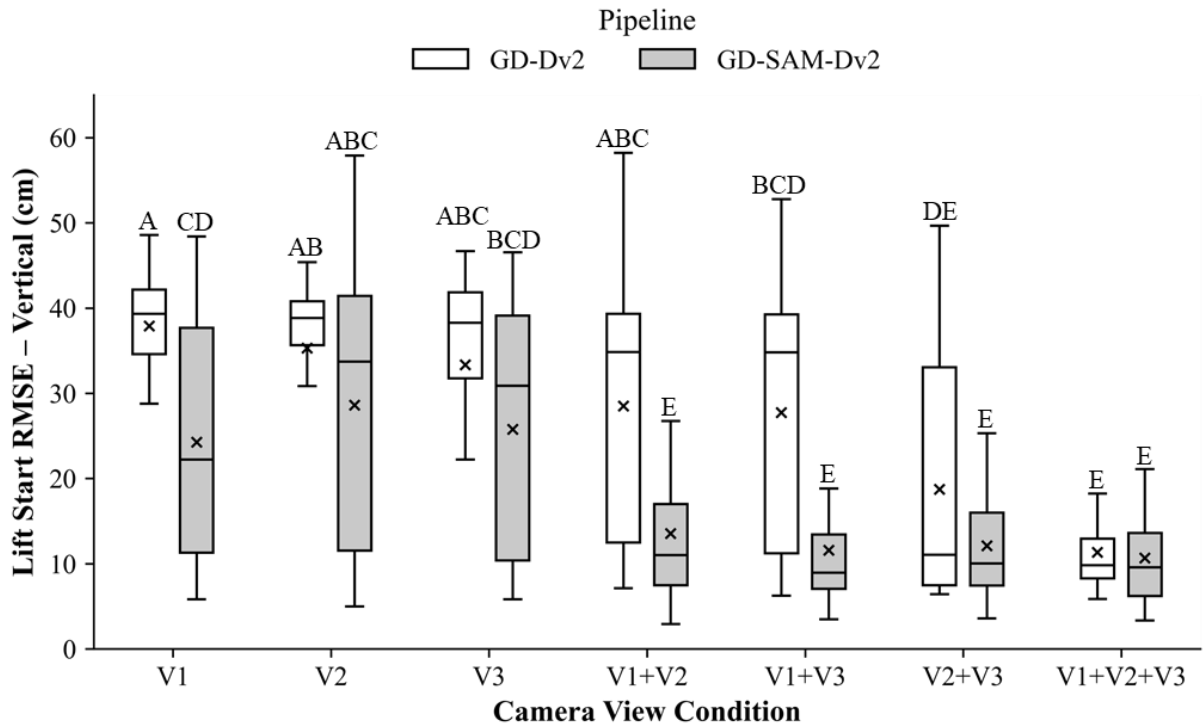

Figure A.4: Significant interaction effect of *Pipeline × Camera View Condition* on RMSE of *V* estimation at the start of the lift.

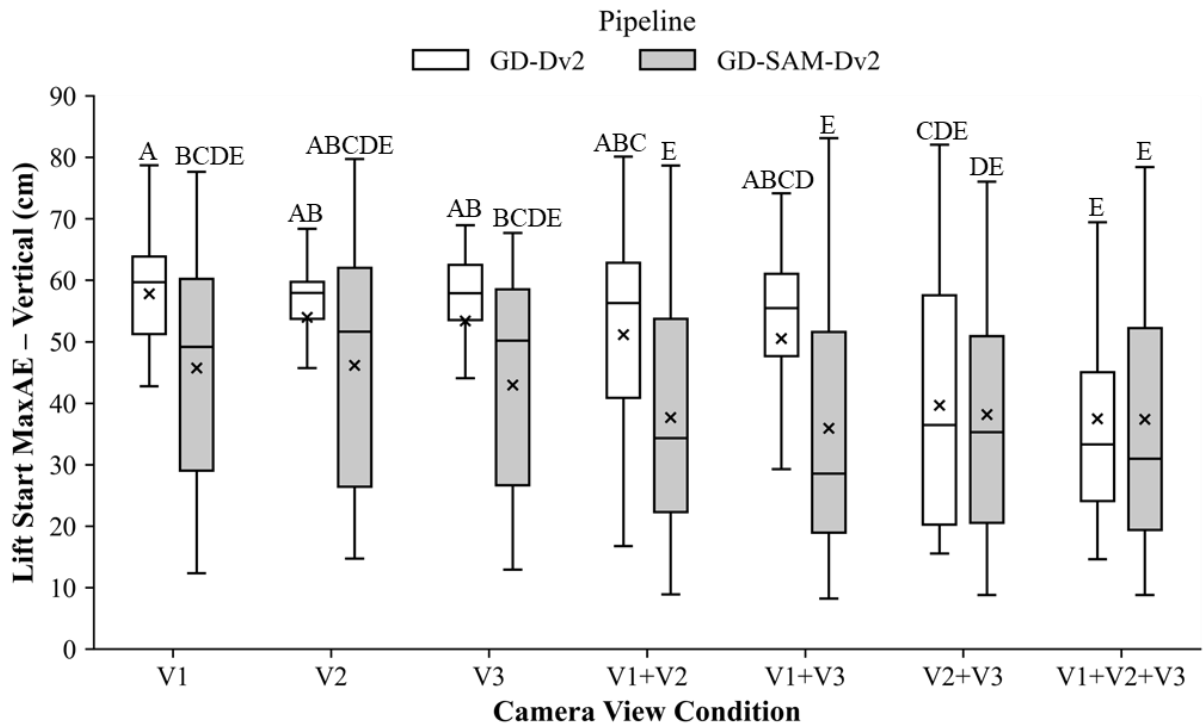

Figure A.5: Significant interaction effect of *Pipeline × Camera View Condition* on MaxAE of *V* estimation at the start of the lift.

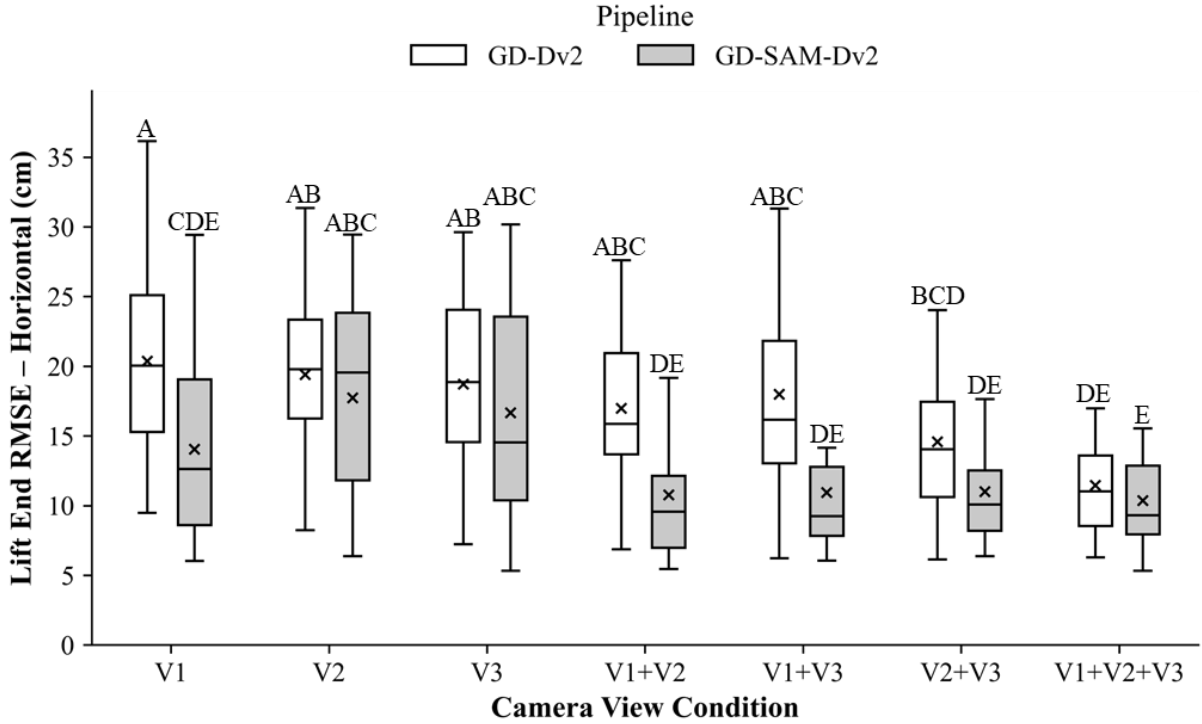

Figure A.6: Significant interaction effect of *Pipeline × Camera View Condition* on RMSE of *H* estimation at the end of the lift.

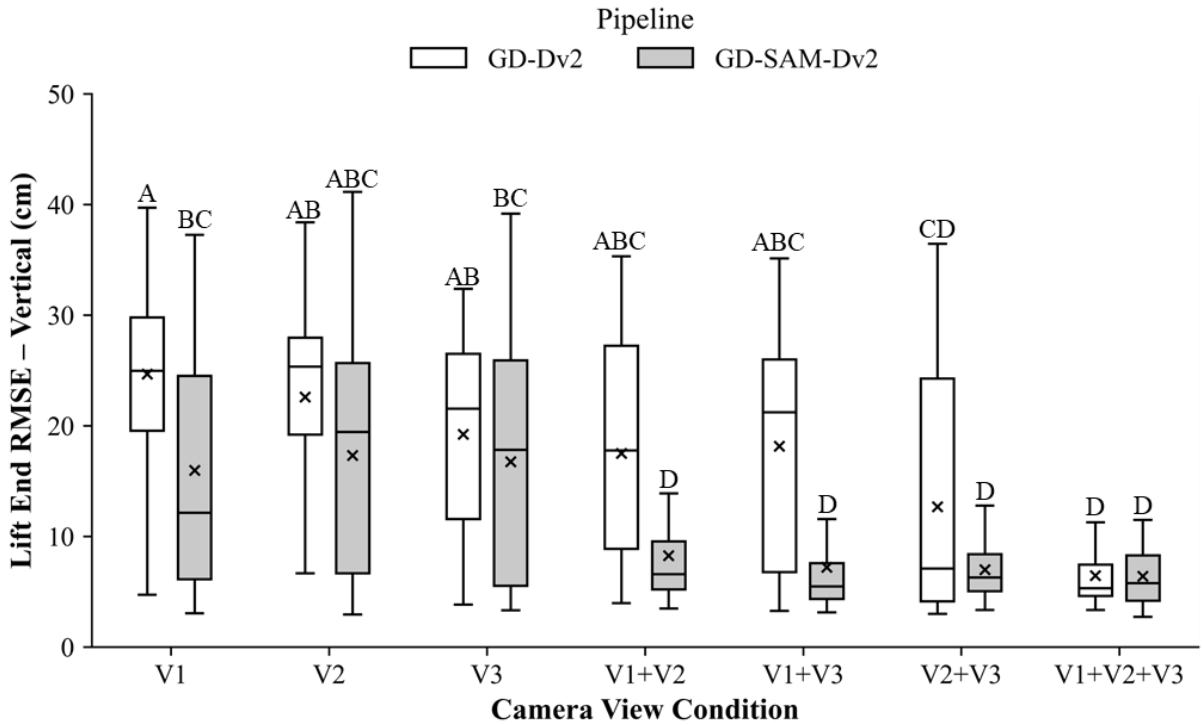

Figure A.7: Significant interaction effect of *Pipeline × Camera View Condition* on RMSE of *V* estimation at the end of the lift.

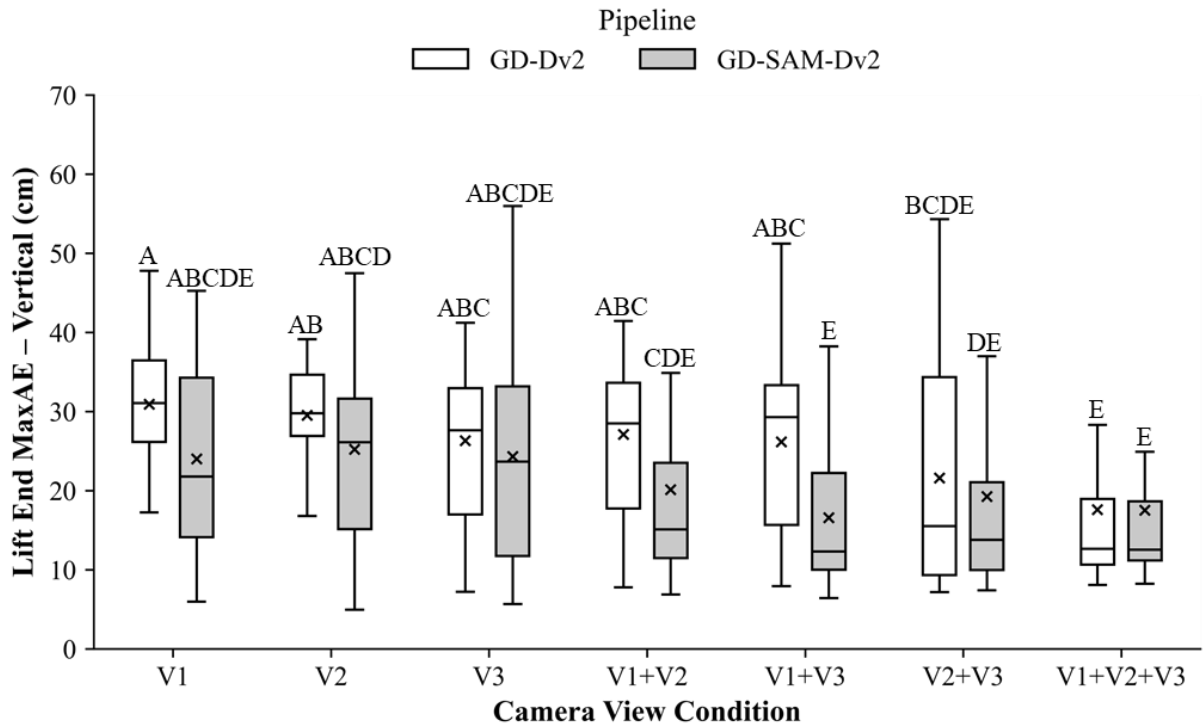

Figure A.8: Significant interaction effect of *Pipeline × Camera View Condition* on MaxAE of *V* estimation at the end of the lift.

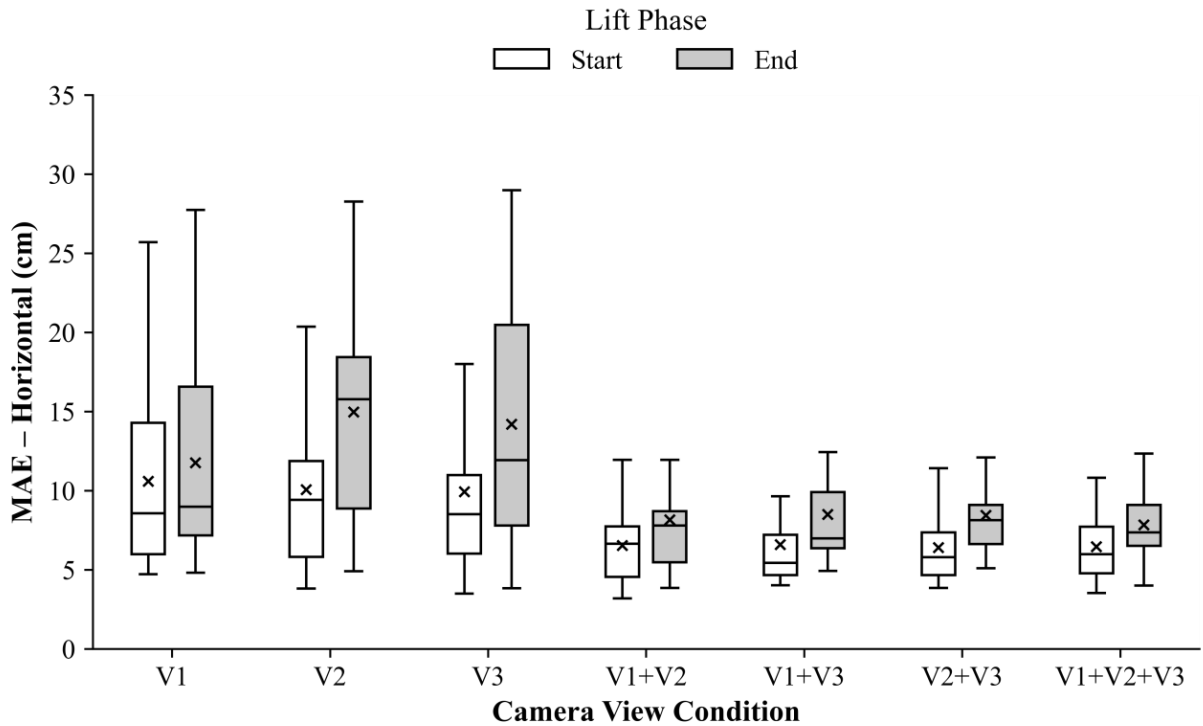
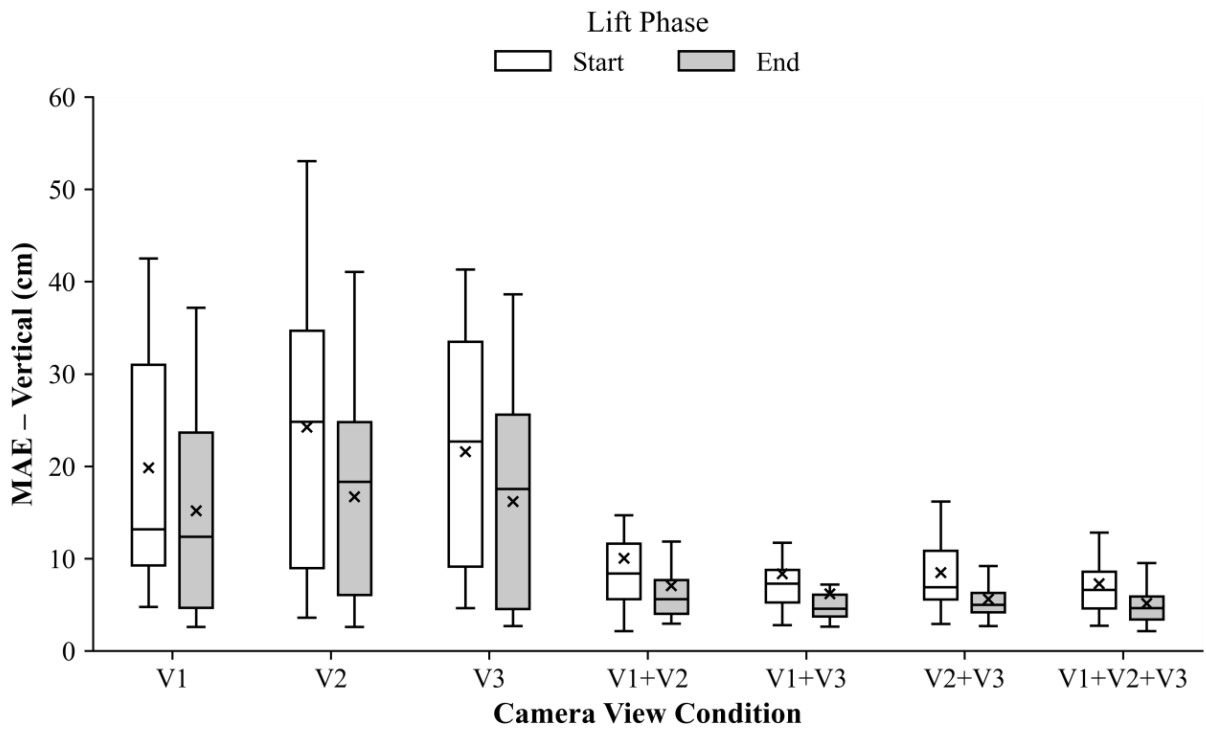

Figure A.9: Comparison of MAE for *H* (top) and *V* (bottom) estimation between the start and end of a lift for the GD–SAM–Dv2 pipeline.